\documentclass[runningheads]{llncs}
\usepackage{graphicx}
\usepackage[colorlinks,allcolors=blue]{hyperref,xcolor}
\usepackage{todonotes}

\begin{document}
\title{Uncertainty Estimation in Deep 2D Echocardiography Segmentation}
\titlerunning{Uncertainty Estimation in Deep 2D Echocardiography Segmentation}
%


\author{Lavsen Dahal \and
Aayush Kafle \and
Bishesh Khanal}
\authorrunning{Dahal et al.}
%
\institute{NepAl Applied Mathematics and Informatics Institute for Research (NAAMII)\\
\email{\{lavsen.dahal,aayush.kafle,bishesh.khanal\}@naamii.org.np}}

\maketitle              
\begin{abstract}

2D echocardiography is the most common imaging modality for cardiovascular diseases. The portability and relatively low-cost nature of Ultrasound (US) enable the US devices needed for performing echocardiography to be made widely available. However, acquiring and interpreting cardiac US images is operator dependent, limiting its use to only places where experts are present. Recently, Deep Learning (DL) has been used in 2D echocardiography for automated view classification, and structure and function assessment. Although these recent works show promise in developing computer-guided acquisition and automated interpretation of echocardiograms, most of these methods do not model and estimate uncertainty which can be important when testing on data coming from a distribution further away from that of the training data. Uncertainty estimates can be beneficial both during the image acquisition phase (by providing real-time feedback to the operator on acquired image's quality), and during automated measurement and interpretation. The performance of uncertainty models and quantification metric may depend on the prediction task and the models being compared. Hence, to gain insight of uncertainty modelling for left ventricular segmentation from US images, we compare three ensembling based uncertainty models quantified using four different metrics (one newly proposed) on state-of-the-art baseline networks using two publicly available echocardiogram datasets. We further demonstrate how uncertainty estimation can be used to automatically reject poor quality images and improve state-of-the-art segmentation results.
 
\end{abstract}
\section{Introduction}
The early symptoms of heart disease such as changes in structure and function of the heart muscle are often detectable by imaging, but screening and longitudinal tracking of such changes are impractical due to the high cost~\cite{zhang2018fully}. Despite the recent advances in handheld portable Ultrasound (US) devices, the challenges remain to improve accessibility as acquiring and interpreting echocardiogram requires expert operators.

The recent success of Deep Learning (DL) has shown great promise in developing automated methods in 2D echocardiography. DL based methods have been used for automated cardiac structure and function assessment, and view classification ~\cite{zhang2018fully,madani2018deep,leclerc2019deep,ouyang2019interpretable}. Left ventricle (LV) volume is one of the essential measures in the cardiac US, which can help accurately estimate ejection fraction~\cite{leclerc2019deep} and would be an integral part of automated echocardiography. Although several methods have been developed for the automated segmentation of LV~\cite{leclerc2019deep,ouyang2019interpretable} in 2D echo images, most of these methods do not include uncertainty estimation. Uncertainty estimation could be beneficial during the image acquisition phase (by providing feedback to the operator on image's quality) \cite{budd2019confident} and during interpretation, providing confidence on the automated measurements obtained from the model to support clinical decision making \cite{nair2020exploring}. For example, ~\cite{leclerc2019deep} manually identify and remove bad quality images for reporting validation dice scores. Being able to automate this process could be useful during the acquisition time or when performing an automatic analysis of the results in a large dataset.

Uncertainty modeling and estimation are being increasingly used in deep learning-based medical imaging applications \cite{leibig2017leveraging,hoebel2019give,budd2019confident,wickstrom2020uncertainty,kim2020automatic,nair2020exploring}. These methods usually produce multiple output predictions for a single input and then measure uncertainty by aggregating information from these outputs. The most popular approach is to approximate Bayesian inference using Monte Carlo dropout~\cite{kendall2015bayesian}, where dropout is used at inference time to sample multiple predictions. Other commonly used approaches to generate multiple samples include using separately trained models~\cite{lakshminarayanan2017simple}, using a selected range of epochs of training called Horizontal Stacked Ensemble (HSE)~\cite{xie2013horizontal}, test time augmentation (TTA) where test input data is augmented and fed multiple times to a single model~\cite{Wang2019,ayhan2018test}, and generative segmentation model with conditional variational autoencoder~\cite{kohl2018probabilistic,baumgartner2019phiseg}. 

Different metrics can be used to estimate uncertainty but the choice of a particular metric is not trivial and careful analysis of various metrics is needed as the best choice may depend on the models being evaluated and the prediction task~\cite{nair2020exploring,ashukha2020pitfalls}. For instance,~\cite{nair2020exploring} provide insightful analysis of various uncertainty metrics (predictive variance, MC sample variance, predictive entropy, and mutual information) for Monte Carlo dropout model for Multiple Sclerosis lesion detection and segmentation.

\textbf{Contribution:} We apply various uncertainty estimation techniques to the convolutional network-based automated LV segmentation of cardiac US images. More specifically: i) In addition to previously used uncertainty measures like variance, entropy and mutual information \cite{nair2020exploring}, we propose probabilistic atlas as an alternative metric (see \ref{sec:methods_metrics}).
ii) We compare, for the first time, the performance of recent methods: MC dropout~\cite{kendall2015bayesian}, TTA~\cite{ayhan2018test}, and relatively less used HSE~\cite{xie2013horizontal}) for measuring uncertainty using four different metrics.
iii) We improve the performance of the current state-of-the-art obtaining higher Dice Similarity Coefficient (DSC) in publicly available test sets when uncertain cases are removed automatically instead of manually removing bad quality images.

\section{Dataset}

Two publicly available datasets in echocardiography - Cardiac Acquisitions for Multi-structure Ultrasound Segmentation (CAMUS)~\cite{leclerc2019deep} and Dynamic-Echonet ~\cite{ouyang2019interpretable}~are used for the experiments. The former dataset has 2D apical four-chamber and two-chamber view sequences of 500 patients. For each sequence, the manual annotation for the End Diastolic(ED) and End Systolic(ES) frames of the left ventricle structures - endocardium, epicardium, and left atrium are provided as the ground-truth for 450 patients. Both 2-chambers and 4-chambers, ED and ES images are shuffled obtaining a total of 1600 images for training, 200 for validation, and 200 for the test. Test set segmentation performance is evaluated on an online platform\footnote{\url{http://camus.creatis.insa-lyon.fr/challenge/\#challenge/5ca20fcb2691fe0a9dac46c8}}.


The Dynamic-Echonet dataset consists of $10,030$ different echocardiography videos with corresponding number of ED and ES frames. For each video, two tracings from experts are provided of both the ED and ES. The US images for ED and ES stages are extracted from the video and frame information, and the ground truth is created from the expert tracings of the left ventricle. The dataset is split into 14956 training, 2552 validation  and 2552 testing images with the same split as \cite{ouyang2019interpretable}.


\section{Methods}

\begin{figure}
\begin{center}
\includegraphics[width=\textwidth]{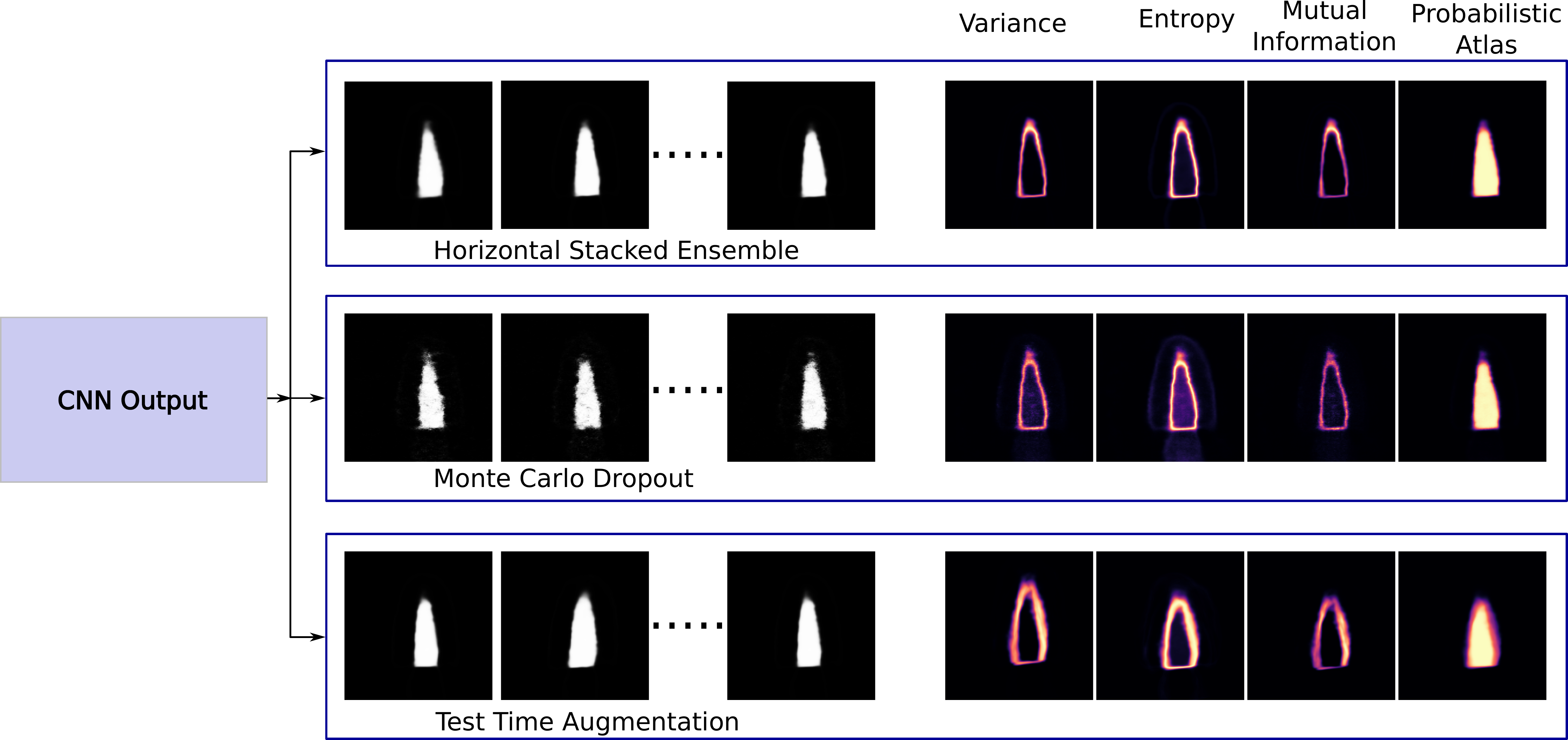}
\caption{Uncertainty modelling and quantification: Each row corresponds to different method to model uncertainty which is quantified by four different metrics shown in the last four columns.}
\label{fig:uncertainty}
\end{center}
\end{figure}

We perform semantic segmentation of echocardiography images and measure test time uncertainty using three different ensembling based models quantified using four different metrics which can be implemented at no additional training cost. Fig.~\ref{fig:uncertainty} shows the uncertainty methods and metrics used. 



\subsection{Semantic Segmentation}
The CAMUS dataset is trained with DeepLab V3+ architecture~\cite{Chen_2018} with Resnet-101 having atrous convolution as the main feature extractors. The Resnet-101~\cite{he2016deep} is pre-trained on ImageNet~\cite{ILSVRC15}. The Deeplab V3+ architecture combines the advantages of both spatial pyramid pooling, and encoder-decoder setup for semantic segmentation and also uses depth-wise separable convolutions. The multi-scale contextual information is captured by spatial pyramid pooling, and the effective receptive field of convolution is controlled by the use of atrous convolution. Moreover, the use of depth-wise separable convolutions reduces computational complexity. The images were resized to 513x513 and fed to the network which is trained with learning rate of 0.007, batch size of 8, and output stride of 16.

The Dynamic-Echonet dataset is trained with EchoNet-Dynamic architecture~\cite{ouyang2019interpretable} with the author's open source implementation\footnote{\url{https://github.com/echonet/dynamic}}. It uses a DeepLabv3~\cite{chen2017rethinking} model with ResNet50~\cite{he2016deep} as the main feature extractor. 

\subsection{Modelling Uncertainty}

\textbf{Monte Carlo Dropout as Bayesian Approximation:}
Supervised training of deep network uses input training images $X$, ground truth labels $Y$ to learn the weights $W$. Since the analytical computation of posterior weight $p(W|X,Y)$ to capture uncertainty from prior distribution of weight$(W)$ is intractable, we use dropout to approximate the distribution of weights~\cite{gal2016uncertainty}. For an input $X^*$, we can now take $T$ samples from the dropout network's segmentation prediction $\hat Y$ to approximate posterior prediction $p(\hat Y|X^*,W)$ as $p(\hat Y|X,Y,X^*) \approx \frac{1}{T}\sum_{t=1}^TP(\hat Y|X^*,W_t)$.


\textbf{Horizontal Stacked Ensemble (HSE) method:}
During training of deep  networks, the validation loss can often oscillate after a certain point of training trajectory without improving any further. However, the training loss may continue to decrease effectively overfitting to the training set. The model tries to fit the distribution of the whole training set, and the validation loss starts oscillating at this stage of training. Inspired from ~\cite{xie2013horizontal,huang2017snapshot}, we save all the models from the epoch where the validation loss stops improving. During inference, we obtain $T$ samples from the softmax outputs of the last layer belonging to the left ventricle class obtained from these saved continuous range of epochs to model the uncertainty.


\textbf{Test Time Augmentation(TTA):}
The augmentation of test images can give multiple output predictions for an image which can be used to model the uncertainty~\cite{Wang2019}. We augment the test image during inference using random rotation in the interval $[-20^0,20^0]$, horizontal flipping, and addition of random Gaussian noise. 



\subsection{Quantifying Uncertainty}

\label{sec:methods_metrics}
We propose and compute a probabilistic atlas based uncertainty measure in addition to other three existing metrics - sample variance, predictive entropy and mutual information~\cite{nair2020exploring,gal2016uncertainty} for which we follow the implementation of~\cite{nair2020exploring}.

\textbf{Probabilistic atlas} retains the information of inter-model variations by averaging the outputs from the last sigmoid or softmax layer for each pixel obtained from different sampling strategies. For each pixel $i$, the $\hat y_{i,avg} = \frac{1}{T}\sum_{t=1}^T\hat y_{i,t}$ is computed to form a probabilistic atlas. It is then thresholded by a value of $h$ (we show results for $h\in$\{0.1,0.5,0.9\}) to obtain binary output segmentations $\hat Y_{h}$. $\hat {DSC}_h$ is computed for all $\hat Y_{h}$ against the predicted segmentation $\hat Y$ ($\hat Y$ is the prediction of the model with lowest validation loss). Here, the test image with the lowest $\hat {DSC}_h$ is selected for downstream rejection based on uncertainty. 


\textbf{Sample variance} is the measure of uncertainty derivation from the variance of $T$ sample outputs from the network for an image $X$. For each pixel $i$, the variance is calculated as $var_i = \frac{1}{T}\sum_{t=1}^T(\hat y_{i,t}-\hat y_{i,avg})^2$.


\textbf{Predictive entropy} of a model is a measure of information carried by the model's predictive density function at each pixel. In our case, entropy is calculated by first computing average prediction ($\hat y_{i,avg}$) for each pixel from all the prediction samples for an input test image ($X$) and then summing the average prediction for each class. Finally, the approximate entropy is given by $H[\hat y_i|x^*_i,X,Y] = -\hat y_{i,avg}\ln\hat y_{i,avg}-(1-\hat y_{i,avg})\ln(1-\hat y_{i,avg})$.



\textbf{Mutual information} between a model's posterior density function and its prediction density function is approximated at each pixel $i$ by computing the difference between predicted entropy and expectation across each sample's entropy i.e. $MI[\hat y_i|x^*_i,X,Y] = H[\hat yi|x^*_i,X,Y]-E[H[\hat y_i|x^*_i,W]]$.




In order to quantify the uncertainty and reject the uncertain images we need to compute the image level uncertainties for all the test images. The proposed metric probabilistic atlas directly provides the image level uncertainty. However, the metrics - sample variance, predictive entropy and mutual information provide the uncertainty in pixel level. Therefore, to propagate the uncertainties to the image level, the log sum of the exponents of the pixel uncertainties is computed followed by the max-min normalization. The highest normalized scores obtained correspond to the most uncertain test cases which are rejected. 

The sample size for HSE methods trained with Deeplabv3+ in CAMUS dataset and Deeplab in Dynamic-Echonet dataset is 100 (model trained for total 300 epochs) and 40 (model trained for total 50 epochs) respectively. The sample sizes are different as the optimum number of samples for HSE is chosen by visually looking at the training graphs when the training is completed for two independent models. For the MC Dropout and TTA, the number of samples is 50 for both the methods.

\section{Results}
{\label{section:results-}}
Table \ref{tab:results} presents the results of uncertainty modeling and quantification, and compares it with the current-state-of-the-art(SoA) baseline models quantitatively. 
The DSC scores reported in the table are the best ones among the various metrics and models used. For CAMUS dataset ED and ES segmentation, the best results were obtained with HSE (metric-mutual information) and TTA (metric-probabilistic atlas) respectively. For Dynamic-Echonet dataset ED and ES segmentation, HSE (metric-probabilistic atlas) and MC Dropout (metric-probabilistic atlas) respectively gave the best results. \cite{leclerc2019deep} reported the results after removing the poor quality images(18.8\%) manually. However, our results are for the test set evaluated online in the CAMUS platform\footnote{\url{http://camus.creatis.insa-lyon.fr/challenge/\#challenge/5ca20fcb2691fe0a9dac46c8}} without any manual intervention based on the proposed uncertainty framework. We obtained higher DSC when filtering 20\% of the most uncertain cases in the Dynamic-Echonet dataset for both ED and ES stages. 

\begin{table}
\centering
\caption{Improved DSC by modelling uncertainty *Results noted in \cite{leclerc2019deep} are for 10 fold cross-validation after manual removal of poor quality images (18.8\%) selected by cardiologists.}
\label{tab:results}
\begin{tabular}{c c c c c c c}
\hline
Test Set & First 20\% & First 40\% & First 60\% & First 80\% & Full-Dataset(100\%) & Current SoA\\
\hline
CAMUS-ED & 0.953 & 0.946 & 0.944 & 0.935 & 0.932 & 0.939*\cite{leclerc2019deep}\\ 
CAMUS-ES & 0.944 & 0.936 & 0.928 & 0.923 & 0.911 & 0.916*\cite{leclerc2019deep}\\ 
\hline
Dynamic-ED & 0.946 & 0.942 & 0.939 & 0.936 & 0.930 & 0.927 \cite{ouyang2019interpretable}\\ 
Dynamic-ES & 0.929 & 0.921 & 0.914 & 0.909 & 0.899 & 0.903 \cite{ouyang2019interpretable}\\ 
\hline
\end{tabular}
\end{table}

\begin{figure}
\begin{center}
\includegraphics[width=\textwidth]{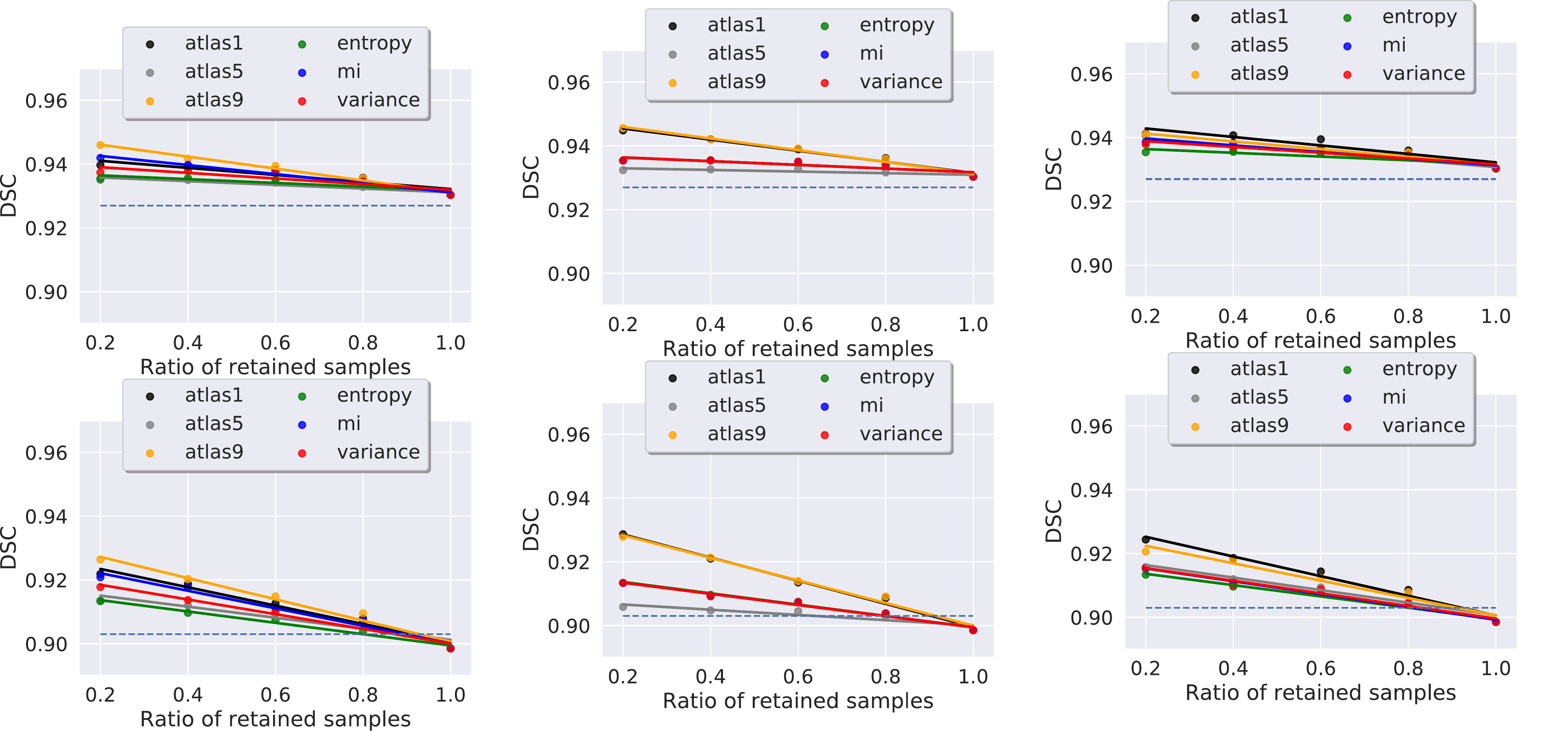}
\caption{Dice Coefficient vs Ratio of Retained Samples for Dynamic-Echonet (in columns from left to right - uncertainty methods : HSE, MC Dropouts, TTA, in rows from top to bottom - ED, ES. The dotted line shows the current SoA DSC)}
\label{fig:graph_dynamic}
\end{center}
\end{figure}

In Fig.~\ref{fig:graph_dynamic}, we present our results for Dynamic-Echonet dataset, comparing all the uncertainty methods. Probabilistic atlas shows the most significant improvement in DSC in all the cases for this dataset. The atlas1, atlas5, and atlas9 correspond to the images obtained with the threshold of 0.1, 0.5, and 0.9 respectively. All three ensembling methods for modeling uncertainty improved DSC as shown in Fig.~\ref{fig:graph_dynamic} which demonstrates that uncertainty estimation at test time helps improve performance of automated segmentation method. The CAMUS dataset showed similar tendency (available in supplementary material). 

In Fig.~\ref{fig:top_bottom_dice}, we visualize and compare the top 2 and bottom 3 images in terms of DSC and the associated uncertainty obtained from HSE method in Dynamic-Echonet test-set. The two uncertainty metrics - variance and mutual information looked similar qualitatively. In two of the bottom three DSCs, we observe higher uncertainty spread over larger area of the image which corresponds to higher uncertainty level for all the metrics except Atlas as seen by the colored uncertainty map. For the probabilistic atlas, the higher intensity in uncertainty map corresponds to lower level of uncertainty. 
And as expected, the bottom 3 images correspond to ES and the Top 2 Images correspond to ED. It is interesting to see that the actual ground-truth does not seem to be consistent for cases with low DSC. However, for the images with top dice scores, all the uncertainty maps show that the model is always highly confident its predictions as seen in Fig.~\ref{fig:top_bottom_dice}.

\begin{figure}
\begin{center}
\includegraphics[width=\textwidth]{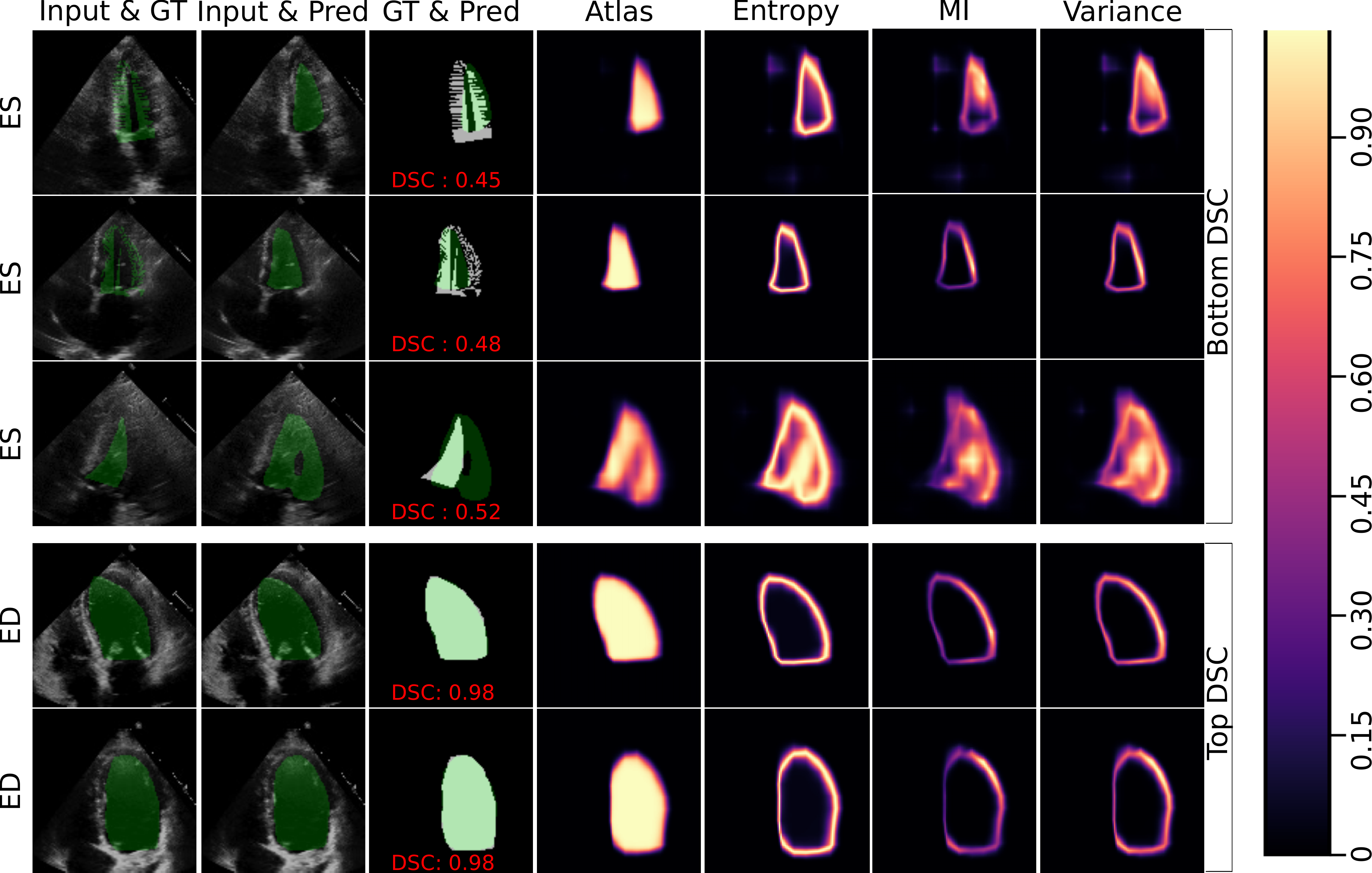}
\caption{Visualization of top and bottom images in terms of DSC for Dynamic Echonet with HSE uncertainty method. The color bar on the right side represents the uncertainty level for the metrics.}
\label{fig:top_bottom_dice}
\end{center}
\end{figure}

\section{Discussion and Conclusion}
We quantified uncertainty for LV segmentation in two recently released publicly available datasets in echocardiography starting from the state-of-the-art baseline results. The experiments show that the ensembling based approaches capturing the uncertainty can improve automated quantification by filtering out difficult or potentially erroneous acquisitions. The average DSC obtained was always higher for ED frames in both the datasets, and most uncertain cases were found in ES frames. We mixed both ED and ES images while training the model to be consistent with the original works, though the distribution of these images is different at least in terms of shape and size. Following this intuition, we trained two distinct models for ED and ES images in the CAMUS dataset. However, the obtained DSC was slightly less, possibly due to the reduction in training dataset size by half. However, training two indepdendent models for the Dynamic-Echonet dataset ED and ES images
could give better results as the training dataset size is quite large (5 times of CAMUS dataset) which we leave as future work. The proposed uncertainty metric probabilistic atlas had the best performance in Dynamic-Echonet dataset in all the cases possibly because it captures the image level uncertainty naturally. Similarly, the probabilistic atlas performed better than other measures in most cases for the CAMUS dataset. The sample size of the test set in the CAMUS dataset was only 50, compared to 1276 in the Dynamic-Echonet dataset. Therefore, the results for the CAMUS dataset could be susceptible to outliers.

We explored ensemble based methods that might be mostly capturing pixel wise variance except the proposed probabilistic atlas based metric. Variational autoencoder based approaches that sample segmentation maps from a latent space might model complex correlation structure in the distribution of plausible segmentations ~\cite{kohl2018probabilistic,baumgartner2019phiseg}. In future, we will explore the impact of using such methods in the current setup. Another interesting line of work is to explore whether our approach could be used to improve ground truth annotations in large database by automatically identifying poorly annotated labels as shown in Fig.~\ref{fig:top_bottom_dice}. Finally, using uncertainty estimates to provide operators feedback needs to run in real time for which the factors such as number of forward inference samples used must be taken into account when choosing a particular method.

\bibliographystyle{splncs04}
\bibliography{bibliography/main.bib}

\section*{Supplementary Material} 

\begin{figure}
\begin{center}
\includegraphics[width=\textwidth]{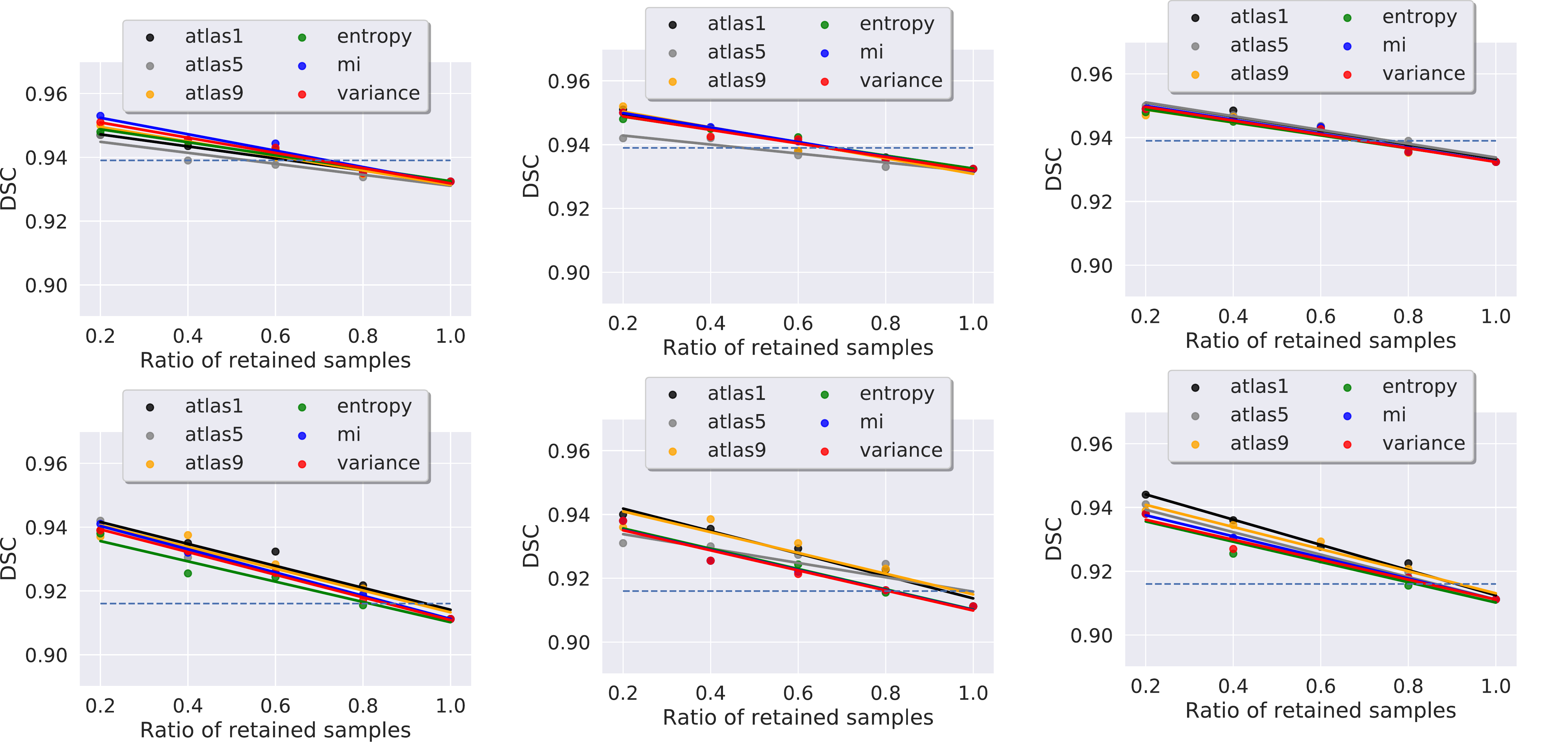}
\caption{Dice Coefficient vs Ratio of Retained Samples for CAMUS (in columns from left to right - uncertainty methods : HSE, MC Dropouts, TTA, in rows from top to bottom - ED, ES. The dotted line shows the current SoA DSC)}
\label{fig:graph_camus_ref}
\end{center}
\end{figure}





\end{document}